\documentclass[sigconf]{acmart}

\copyrightyear{2025}
\acmYear{2025}
\setcopyright{acmlicensed}
\acmConference[CIKM '25] {Proceedings of the 34th ACM International Conference on Information and Knowledge Management}{ November 10--14, 2025}{Seoul, Republic of Korea.}
\acmBooktitle{Proceedings of the 34th ACM International Conference on Information and Knowledge Management (CIKM '25), November 10--14, 2025, Seoul, Republic of Korea}
\acmISBN{979-8-4007-2040-6/2025/11}
\acmDOI{10.1145/XXXXXX.XXXXXX}

\usepackage{graphicx}
\usepackage{multirow}
\usepackage[caption=false,font=footnotesize,labelformat=parens,labelsep=space]{subfig}

\begin{document}

\title{\textit{V-Agent}: An Interactive Video Search System Using Vision-Language Models}

\author{SunYoung Park}
\authornote{Equal Contribution.}
\affiliation{%
  \institution{NC AI}
  \city{Seongnam-si}
  \state{Gyeonggi-do}
  \country{Republic of Korea}
}
\email{sun0park@ncsoft.com}

\author{Jong-Hyeon Lee}
\affiliation{%
  \institution{NC AI}
  \city{Seongnam-si}
  \state{Gyeonggi-do}
  \country{Republic of Korea}
}
\authornotemark[1]
\email{leejh1230@ncsoft.com}

\author{Youngjune Kim}
\authornotemark[1]
\authornote{The author completed this work while working at NC AI.}
\affiliation{%
  \institution{Kakao}
  \city{Seongnam-si}
  \state{Gyeonggi-do}
  \country{Republic of Korea}}
\email{daniel.log@kakaocorp.com}

\author{Daegyu Sung}
\authornote{The author completed this work while interning at NC AI.}
\affiliation{%
  \institution{Korea Advanced Institute of Science and Technology}
  \city{Daejeon}
  \country{Republic of Korea}}
\email{sbigstar0310@kaist.ac.kr}

\author{Younghyun Yu}
\affiliation{%
  \institution{NC AI}
  \city{Seongnam-si}
  \state{Gyeonggi-do}
  \country{Republic of Korea}
}
\email{zrohyun@ncsoft.com}

\author{Young-rok Cha}
\affiliation{%
  \institution{NC AI}
  \city{Seongnam-si}
  \state{Gyeonggi-do}
  \country{Republic of Korea}
}
\email{jaycha@ncsoft.com}

\author{Jeongho Ju}
\affiliation{%
  \institution{NC AI}
  \city{Seongnam-si}
  \state{Gyeonggi-do}
  \country{Republic of Korea}
}
\email{jeongho@ncsoft.com}

\renewcommand{\shortauthors}{Park et al.}

\begin{abstract}
We introduce \textit{V-Agent}, a novel multi-agent platform designed for advanced video search and interactive user-system conversations. By fine-tuning a vision-language model (VLM) with a small video preference dataset and enhancing it with a retrieval vector from an image-text retrieval model, we overcome the limitations of traditional text-based retrieval systems in multimodal scenarios. The VLM-based retrieval model independently embeds video frames and audio transcriptions from an automatic speech recognition (ASR) module into a shared multimodal representation space, enabling \textit{V-Agent} to interpret both visual and spoken content for context-aware video search. This system consists of three agents—a routing agent, a search agent, and a chat agent—that work collaboratively to address user intents by refining search outputs and communicating with users. The search agent utilizes the VLM-based retrieval model together with an additional re-ranking module to further enhance video retrieval quality. Our proposed framework demonstrates state-of-the-art zero-shot performance on the MultiVENT 2.0 benchmark, highlighting its potential for both academic research and real-world applications. The retrieval model and demo videos are available at \url{https://huggingface.co/NCSOFT/multimodal-embedding}.
\end{abstract}



\maketitle

\begin{figure}[!h]
    \centering
\includegraphics[width=0.44\textwidth]{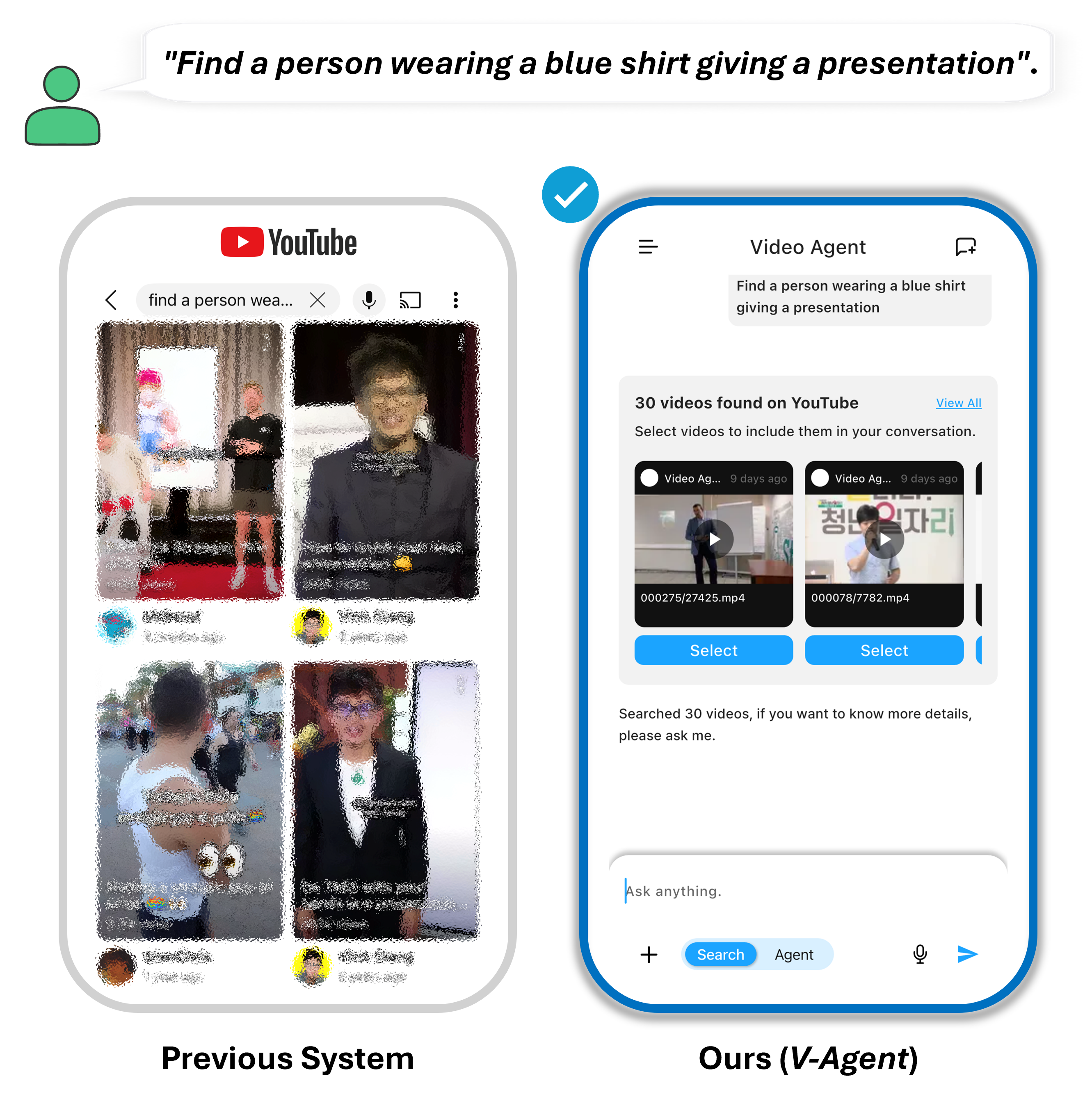}
    \caption{Comparison of search results for the query "Find a person wearing a blue shirt giving a presentation". YouTube returns unrelated videos, whereas our system retrieves semantically relevant results by analyzing visual content.}
    \label{fig:youtube}
\end{figure}

\section{Introduction}
With the increasing popularity of AI assistants such as ChatGPT\footnote{\url{https://chatgpt.com/}} and Perplexity\footnote{\url{https://www.perplexity.ai/}}, search systems are evolving to support multiple modalities beyond traditional text retrieval. The conventional text-based retrieval pipeline \cite{robertson1994okapi} faces significant limitations in multimodal scenarios, as it restricts both queries and candidates to textual formats. Even on YouTube, where millions of users are actively engaged, the search system still relies on metadata such as titles, tags, and descriptions, rather than analyzing the actual video content. Current AI-driven systems handle text and images effectively, but many lack conversational support for referencing and providing comprehensive insights from multiple videos.

Recent research on vision-language models (VLMs) has broadened their scope to videos \cite{wang2024qwen2vlenhancingvisionlanguagemodels, wang2024internvideo2scalingfoundationmodels, chen2023valor, chen2024vast}, and many of these models are showing remarkable performances on numerous video understanding benchmarks. However, prior studies tend to concentrate on VLM training methods and benchmark performance, often not fully exploring the potential benefits of integrating the models into systematic frameworks for real-world applications. 

We introduce \textbf{\textit{V-Agent}}, a video-specialized platform that allows users to search for relevant videos and have interactive conversations. The system employs three distinct agents: a routing agent, a search agent, and a chat agent. They collaborate to process user search requests and generate comprehensive results based on retrieved content.

For effective video retrieval in \textit{V-Agent}, we propose an efficient fine-tuning method that adapts VLMs into video-text retrieval models. First, we fine-tune an instruction-tuned VLM on a small-scale video preference dataset, training the model to rank positive answer above negative responses. Next, we derive a retrieval vector by subtracting the weights of the original instruction-tuned VLM from those of a well-trained image-text retrieval model, and add this vector to the fine-tuned model. Incorporating the retrieval vector improves the model’s general capacity in vision-based retrieval tasks and helps address challenges associated with limited video training data. The VLM-based video-text retrieval model is utilized in both the search agent and data indexing processes to generate visual and textual embeddings. 

During the indexing process, video frames and transcribed audio texts are embedded separately using the retrieval model. At query time, upon receiving a user query, the routing agent determines whether video retrieval is necessary. If so, the search agent employs the VLM-based video-text retrieval model to retrieve the most relevant video frames and transcribed texts. The top-k results from each modality are then fused using a scoring fusion approach, and refined by a re-ranking module. Given the final ranked list of videos, the chat agent analyzes the user query and generates interactive responses related to the videos.
 
In this paper, we investigate how our framework enhances overall quality and user experience in video search through extensive experiments under various settings. In addition to providing a user-centric interface for a video-oriented chat platform using Flutter\footnote{\url{https://flutter.dev/}}, \textit{V-Agent} achieves superior performance on the MultiVENT 2.0 benchmark \cite{kriz2025multivent20massivemultilingual}. Experimental results and use cases further highlight the effectiveness of our approach in both academic and real-world contexts.

\section{Backgrounds}

\textbf{Multimodal Retrieval.} The advancement of multimodal retrieval has been driven by recent progress in language-image encoder models \cite{radford2021learningtransferablevisualmodels, jia2021scalingvisualvisionlanguagerepresentation, EVA-CLIP}. Previous studies have explored adaptation of model architectures and training strategies to incorporate videos into encoder-based models \cite{Luo2021CLIP4Clip, fang2021clip2video, xue2023clipvipadaptingpretrainedimagetext, jin2022expectationmaximizationcontrastivelearningcompact, zhu2023languagebind}. Yet, they still have weaknesses in handling complex queries and the multimodal aspects in videos, particularly audio. The emergence of VLMs has led to significant breakthroughs in text-image retrieval \cite{lin2024nvmmembed, zhang2024gme, zhou2024megapairs, jiang2025vlm2vectrainingvisionlanguagemodels}, as they use large language models (LLMs) capable of understanding intricate text and instructions. Nonetheless, most VLM-based retrieval methods are currently limited to single-image settings and require enormous amounts of training data.

For video foundation models, the training process often demands significantly more resources than that of image foundation models \cite{wang2024internvideo2scalingfoundationmodels, chen2023valor, chen2024vast}. While these models excel in simple video-caption retrieval tasks, their performance tends to decline in more complex video retrieval scenarios. To overcome the challenge, MMMORRF \cite{samuel2025mmmorrfmultimodalmultilingualmodularized} introduces a modality-aware fusion search engine using frame information, optical character recognition (OCR), and automatic speech recognition (ASR). However, the integration of VLMs into video retrieval frameworks remains underexplored.

\textbf{LLM-based Re-ranking.} Since LLMs demonstrate excellent language generalization ability, there have been approaches to incorporate them in document re-ranking tasks \cite{Sun2023IsCG,TourRank, qin-etal-2024-large}. Unlike traditional neural ranking models which require training the models on in-domain large-scale retrieval dataset, LLMs with appropriate re-ranking instructions achieve remarkable performance on text retrieval tasks even in zero-shot settings. These recent findings suggest that employing LLM-based re-ranking can be an effective strategy for real-world applications.

\textbf{Agent Systems.} Recent AI systems have progressed toward multi-agent architectures that combine multiple specialized modules to perform complex tasks \cite{guo2024largelanguagemodelbased}. In particular, LLM-based agents can dynamically select appropriate modules or control task workflows according to task complexity. Prior work demonstrates that agent-based systems integrating retrieval, re-ranking, and conversational interfaces improve information accessibility and increase user satisfaction \cite{schneider2024engineeringconversationalsearchsystems, zhang2025surveylargelanguagemodel, feng2023synergisticinterplaysearchlarge}. 
Following this emerging trend, this study proposes an integrated approach utilizing agents for video search and conversational response systems.

\section{\textit{V-Agent}}
\subsection{Video-Text Retrieval Model}
\label{subsection:vtrm}
For \textit{V-Agent}, we transform Qwen2-VL-7B-Instruct\footnote{\url{https://huggingface.co/Qwen/Qwen2-VL-7B-Instruct}} \cite{wang2024qwen2vlenhancingvisionlanguagemodels} into an effective video retrieval model. The video-text retrieval model $M_R$ is used during both indexing and search to embed videos and texts. 

\begin{figure*}[t!]
    \centering
    \includegraphics[width=\textwidth]{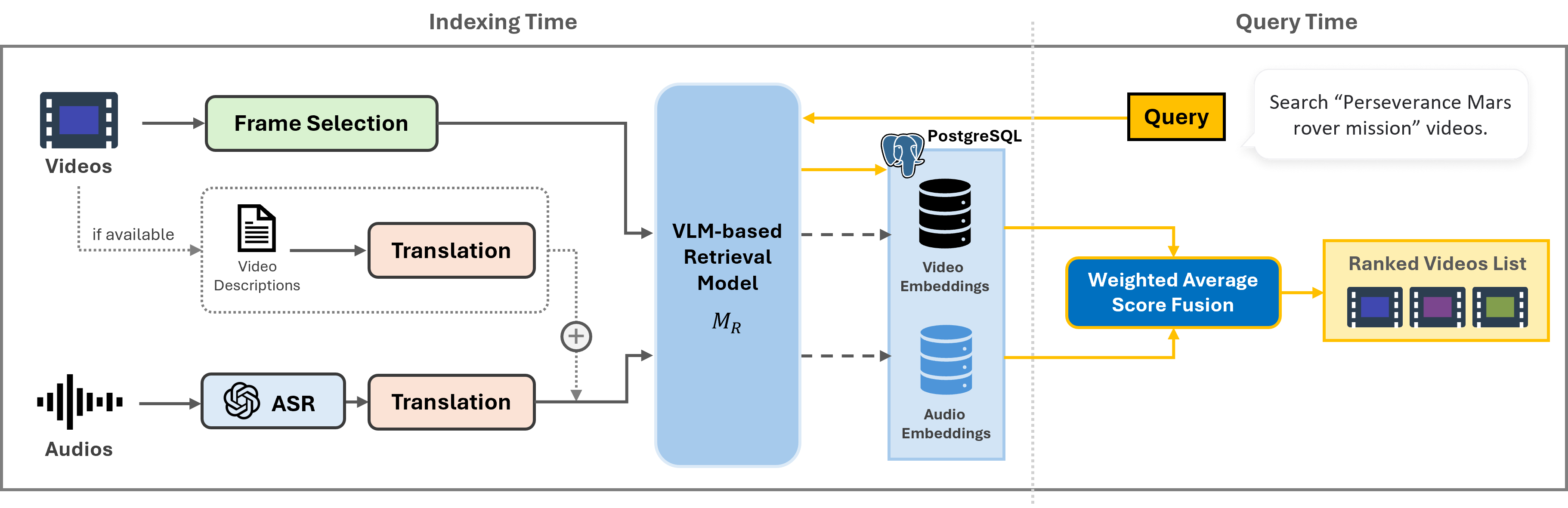}
    \vspace{-6mm}
    \caption{Video Search Pipeline. The VLM-based retrieval model ($M_R$) embeds videos, audio, and video descriptions during both indexing and query processing. Upon receiving a query, the system retrieves the most relevant videos and transcriptions. The results are then combined using a weighted average score fusion method.}
    \label{fig:vsp}
    \vspace{-1mm}
\end{figure*}

\begin{figure}[t!]
    \centering
\includegraphics[width=0.47\textwidth]{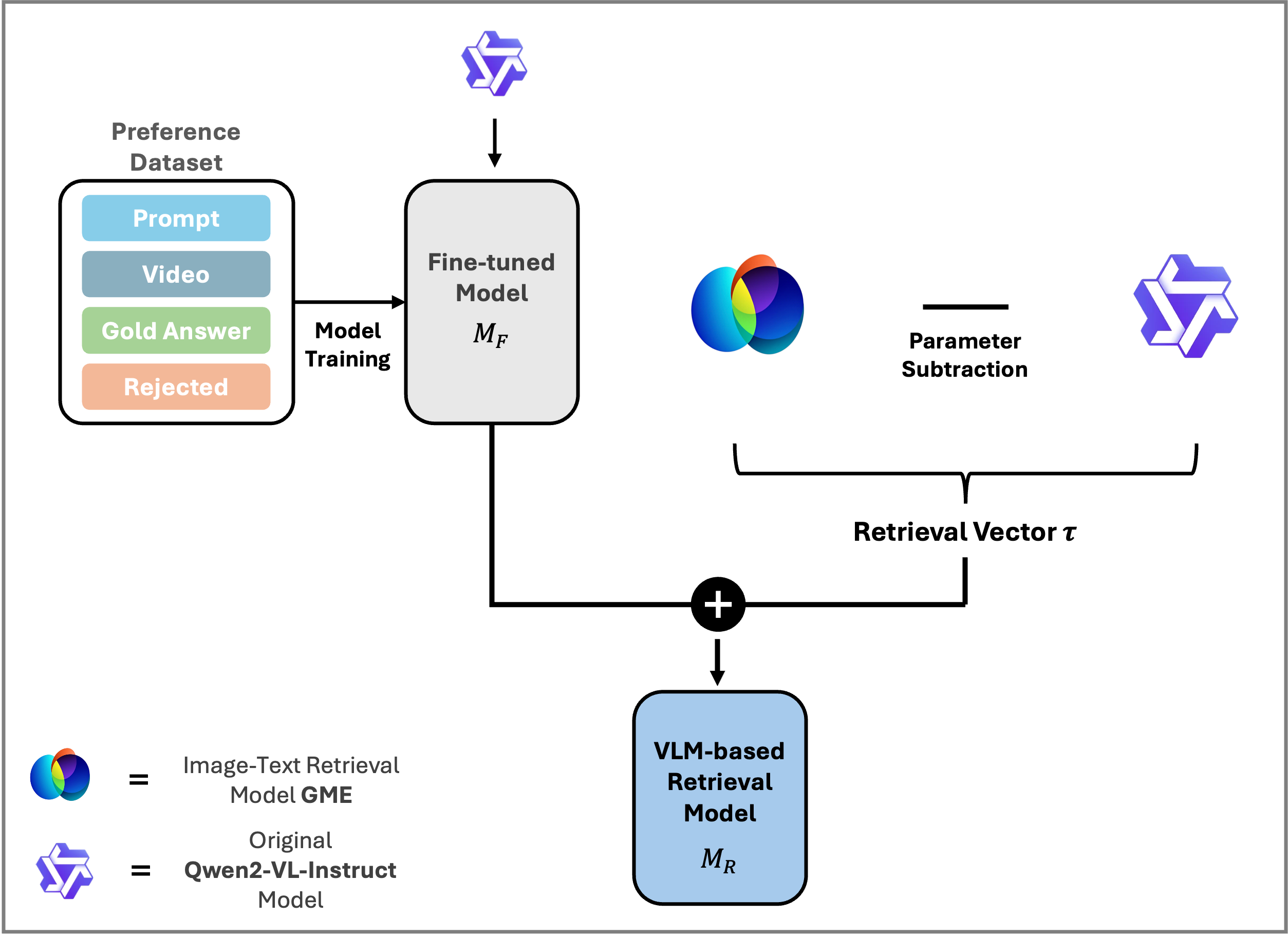}
    \caption{Video-Text Retrieval Model Overview. We fine-tune a Model using the video preference dataset, and then add a retrieval vector $\tau$ to the model to enhance the overall vision-text retrieval capability. $\tau$ is computed as the parameter difference between an image-text retrieval model (GME) and the original Qwen2-VL-Instruct model.}
    \label{fig:mtp}
    
\end{figure}

\subsubsection{Fine-tuning} 
\label{subsub:finetuning}
To efficiently fine-tune the model, we utilize ShareGPTVideo’s $17k$
video preference dataset\footnote{\url{https://huggingface.co/datasets/ShareGPTVideo/train_video_and_instruction}}\cite{zhang2024direct}, which includes prompts, videos, gold answers, and chosen-rejected text pairs. We treat the prompts and videos as queries, and the rejected responses as hard-negatives for the gold answers. The inputs $x_i$ (system and user prompts concatenated with videos), $x_i^+$ (the gold answer for the prompt), and $x_i^-$ (a rejected response for the prompt) are embedded as $h_i$, $h_i^+$, and $h_i^-$, respectively, using the last hidden states of the EOS tokens. Each query $x_i$ is trained with in-batch negatives as well as one hard negative using the InfoNCE loss \cite{oord2019representationlearningcontrastivepredictive}. We adapt code from LLaMA-Factory \cite{zheng2024llamafactory} to train a video-text retrieval model. The model is fully fine-tuned for two epochs on 8 A100 GPUs with a batch size of 8, requiring only a few hours for training. 


\subsubsection{Adding Retrieval Vector} To compensate for the insufficiency of training instances and enhance the generalization ability of the fine-tuned model $M_F$ in \ref{subsub:finetuning}, we compute a retrieval vector $\tau$ by subtracting the weights of the original Qwen2-VL-7B-Instruct model from those of GME\footnote{\url{https://huggingface.co/Alibaba-NLP/gme-Qwen2-VL-7B-Instruct}} \cite{zhang2024gme}, a Qwen2-VL based image-text retrieval model. This approach is inspired by \textit{Chat Vector} \cite{huang-etal-2024-chat}, which is a method to equip pre-trained language models with chat capabilities in new languages by adding a vector obtained from the weight difference between a base model and its chat-optimized counterpart. 

Let $\theta_{GME}$, $\theta_{Qwen}$, and $\theta_{M_F}$ be the weights of GME, Qwen, and the fine-tuned model $M_F$, respectively. We define the retrieval vector $\tau$ as follows:

\begin{equation}
    \tau = \theta_{GME} - \theta_{Qwen}
\end{equation}
\begin{equation}
    \theta_{M_R} = \theta_{M_F} + \tau
\end{equation}

where $\theta_{M_R}$ are the weights for the final video-text retrieval model $M_R$.
The retrieval vector $\tau$ enhances the model's vision-text alignment, which is fundamental for video retrieval tasks where visual frames must be matched with textual queries, resulting in improved video search performance.

\begin{figure*}[!t]
    \centering
    \includegraphics[width=\textwidth]{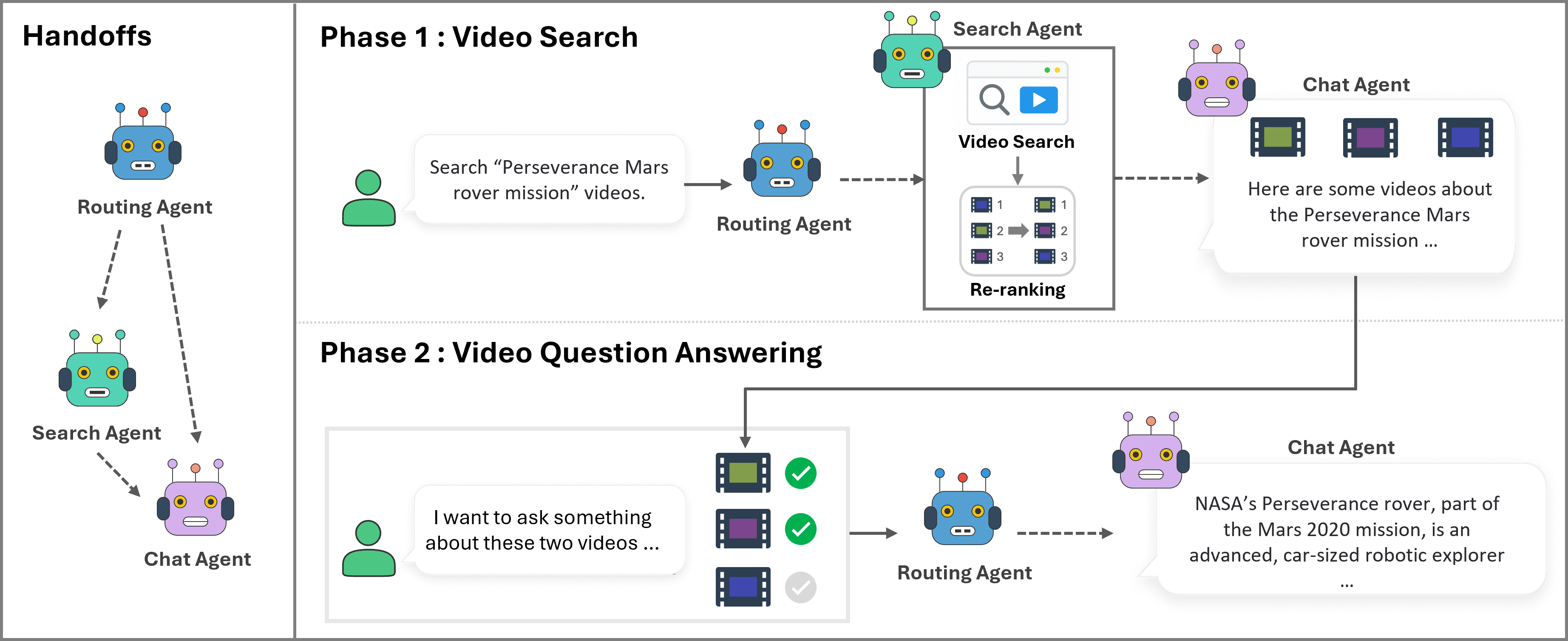}
    \caption{\textit{V-Agent} Pipeline. When a user submits a query intending to search for videos, the routing agent forwards the request to the search agent. The search agent employs the VLM-based video-text retrieval model to retrieve the most relevant video frames and transcribed texts. The results from each modality are then fused using a scoring fusion approach, and the top-$k$ video list is refined by an LLM re-ranking module. Once relevant results are retrieved, users can freely select videos of interest and ask questions about them. Otherwise, the routing agent directly calls the chat agent.}
    \vspace{-1mm}
    \label{fig:vap}
\end{figure*}

\subsection{Video Search}
Figure \ref{fig:vsp} illustrates the indexing and retrieval processes of the video search pipeline in \textit{V-Agent}.

\subsubsection{Indexing Time}
Considering that employing only a video retrieval model would solely capture the visual aspects of videos, we include an ASR model to transcribe the audio content. Given a video, audio transcription is performed using OpenAI's \texttt{Whisper}\footnote{\url{https://huggingface.co/openai/whisper-large-v3-turbo}} \cite{radford2022robustspeechrecognitionlargescale}. If a video description is available, it is concatenated with the transcription to provide additional context. In cases where it is not in English, the combined text is translated into English using \texttt{gpt-4o-mini}\footnote{\label{gpt4ominifootnote}\url{https://platform.openai.com/docs/models/gpt-4o-mini}}. For each video, we extract 48 frames at uniform intervals. The selected video frames and corresponding transcriptions are then simultaneously embedded by the $M_R$ model, which is capable of processing both textual and visual inputs. The vision and audio transcription embeddings are indexed using \texttt{pgvector}\footnote{\url{https://github.com/pgvector/pgvector}} with HNSW settings of $m=16$ and \textit{ef\_construction }$=200$.

\subsubsection{Query Time}
\label{querytime}
At query time, the retrieval model $M_R$ encodes the input query $q_\mathrm{user}$ into an embedding representation, and retrieves the most relevant video frames and audio transcriptions from the database. We calculate the importance score of each video $i$ by fusing the inner-product scores of its frames and transcriptions, and then rank the videos based on these scores.
\begin{equation}
    e_q = M_R(q_\mathrm{user})
\end{equation}
\begin{equation}
    e_f = M_R(f_1, f_2, ..., f_N)
\end{equation}
\begin{equation}
    e_a = M_R(\texttt{audio\_transcription})
\end{equation}
\begin{equation}
score = \alpha\cdot \langle e_f, e_q \rangle + (1 - \alpha) \cdot  \langle e_a, e_q \rangle
\end{equation}
where $N$ denotes the number of frames we select for each video. $e_q$, $e_f$, and $e_a$ represent embeddings for the input query, frames, and audio transcription, respectively.  We set $\alpha$ to 0.5 in this work. 

\subsection{Agent Pipeline}
Three agents–the routing agent, search agent, and chat agent–cooperate to respond to a user request through two phases of video search and video question answering. We employ OpenAI Agents SDK\footnote{\url{https://openai.github.io/openai-agents-python/}} to call APIs for the agents. The api versions of each agent are listed in Table~\ref{table:agent_api}.

\begin{table}[!h]
\small
\caption{API Version of Agents.}
\vspace{-2mm}
\resizebox{0.6\columnwidth}{!}{%
\begin{tabular}{l|c}
\hline \hline
 & \textbf{API Version} \\ \hline
$A_{\mathrm{Routing}}$ & \texttt{gpt-4.1-mini}\footnote{\url{https://platform.openai.com/docs/models/gpt-4.1-mini}} \\ 
$A_{\mathrm{Search}}$, $A_{\mathrm{Chat}}$ & \texttt{gpt-4o}\footnote{\label{gpt4ofootnote}\url{https://platform.openai.com/docs/models/gpt-4o}} \\ 
Reranker (\texttt{rerank}) & \texttt{gpt-4o-mini}\footref{gpt4ominifootnote}  \\ 
\hline \hline        
\end{tabular}
}
\vspace{-2mm}
\label{table:agent_api}
\end{table}

\subsubsection{Routing Agent} 
\label{sec:routing}

The routing agent $A_{\mathrm{Routing}}$ determines whether a given user query $q_{\mathrm{user}}$ should be processed by the search agent or directly by the chat agent. If the query requires video retrieval, the request is handed off to the search agent $A_{\mathrm{Search}}$, and otherwise it is passed to the chat agent $A_{\mathrm{Chat}}$ for direct conversation.

\begin{equation}
A_{\mathrm{Routing}}(q_{\mathrm{user}}) \rightarrow
\begin{cases}
    A_{\mathrm{Search}}(q_{\mathrm{user}}) & \text{if retrieval is required} \\
    A_{\mathrm{Chat}}(q_{\mathrm{user}}) & \text{otherwise}
\end{cases}
\end{equation}
where the arrow ($\rightarrow$) indicates the handoff from the routing agent to the search agent or the chat agent.

\subsubsection{Search Agent}
\label{sec:search}

The search agent $A_{\mathrm{Search}}$ follows a two-step pipeline. First, the \texttt{search\_video} tool retrieves the top-$k$ candidate videos $V_{1:k}$ using the retrieval model $M_R$. Each retrieved video is represented as a tuple of audio transcription $a_i$ and textual description $d_i$: 
\begin{equation}
\begin{aligned}
V_{1:k} = \mathtt{search\_video}(q_{\mathrm{user}}) \\ = \{(a_1; d_1), \ldots, (a_k; d_k)\} 
\end{aligned}
\end{equation}

Next, the \texttt{rerank} tool reorders $V_{1:k}$ based on the user query $q_{\mathrm{user}}$. The reranker is an LLM-based module guided by a specific prompt $p_{\mathrm{rerank}}$. The reranked results $\hat{V}_{1:k}$ are then passed through a handoff to the chat agent $A_{\mathrm{Chat}}$. We set $k = 10$.

\begin{equation}
\hat{V}_{1:k} = \mathtt{rerank}(p_{\mathrm{rerank}}, q_{\mathrm{user}}, V_{1:k})
\end{equation}
\begin{equation}
A_{\mathrm{Search}}(q_{\mathrm{user}}) = \hat{V}_{1:k}, \quad 
A_{\mathrm{Search}} \rightarrow A_{\mathrm{Chat}}
\end{equation}

\subsubsection{Chat Agent}
\label{sec:chat}

The chat agent $A_{\mathrm{Chat}}$ operates in two scenarios. If it receives $l$ videos as input, the chat agent generates responses grounded in the provided video contents. $l$ denotes the number of input videos that are either retrieved by the search agent (Phase~1 in Figure~\ref{fig:vap}) or selected by the user (Phase~2 in Figure~\ref{fig:vap}). When videos are not given, it engages in open-domain conversation without video context.

\begin{equation}
output =
\begin{cases}
    A_{\mathrm{Chat}}(q_{\mathrm{user}}, \hat{V}_{1:l}) & \quad  1 \leq l \leq k \\
    A_{\mathrm{Chat}}(q_{\mathrm{user}}) & \quad l = 0
\end{cases}
\end{equation}


\section{Evaluation Settings}
\subsection{MSR-VTT}
For evaluating the video-text retrieval model $M_R$ mentioned in ~\ref{subsection:vtrm}, we use the widely adopted MSR-VTT benchmark \cite{xu2016msr-vtt}. MSR-VTT is an open-domain dataset containing 10,000 videos, each annotated with five distinct captions, for a total of 200K captions. Following prior work \cite{jsfusion}, we evaluate on the standard 1K test split. The list of the tested models is as follows:

\begin{itemize}
    \item \textbf{Qwen2-VL-7B-Instruct} \cite{wang2024qwen2vlenhancingvisionlanguagemodels} is an instruction-tuned multimodal large language model that understands and reasons over text, images, and long videos.
    \item \textbf{GME-Qwen2-VL-7B-Instruct} \cite{zhang2024gme} is a model designed for universal multimodal embeddings based on Qwen2-VL-7B-Instruct. GME currently accepts only single images; therefore, we experiment with both mean-pooling single-image embeddings across frames and feeding multiple frames simultaneously for multi-image inference.
    \item \textbf{LamRA} \cite{liu2025lamra} is a VLM with advanced retrieval and reranking capabilities by integrating lightweight LoRA modules, enabling them to handle diverse retrieval tasks through a two-stage training strategy and joint reranking techniques.
    \item \textbf{InternVideo2-6B} \cite{wang2024internvideo2scalingfoundationmodels} is a large video foundation model for understanding, describing, and reasoning over video content.
    \item \textbf{Ours ($M_F$ and $M_R$)} are the models described in Section~\ref{subsection:vtrm}. $M_F$ is the fine-tuned model trained on the video preference dataset, and $M_R$ is the final retrieval model obtained by adding the retrieval vector $\tau$ to $M_F$.
\end{itemize}

\subsection{MultiVENT 2.0}
We evaluate our pipeline on MultiVENT 2.0 \cite{kriz2025multivent20massivemultilingual}, a comprehensive multilingual benchmark for event-centric video retrieval. While MSR-VTT is a relatively simple benchmark for matching videos with corresponding captions, MultiVENT 2.0 comprises 3,900 queries constructed from visual content, audio, embedded text, and metadata from videos in six languages (Arabic, Chinese, English, Korean, Russian, and Spanish). For evaluation, we use only the MultiVENT-TEST split, which contains 2,545 queries and 109,800 videos. We reference the official MULTIVENT 2.0 leaderboard\footnote{\url{https://eval.ai/web/challenges/challenge-page/2507/evaluation}}
 to conduct a comparative analysis of the models listed below against ours.

\begin{itemize}
    \item \textbf{InternVideo2-6B} \cite{wang2024internvideo2scalingfoundationmodels}
    \item \textbf{VAST} \cite{chen2024vast} is a foundation model that processes vision, audio, subtitles, and text from videos using omni-modality representations.
    \item \textbf{CLIP} \cite{radford2021learningtransferablevisualmodels} is a model that learns joint image–text representations for zero-shot classification and retrieval.
    \item \textbf{LanguageBind} \cite{zhu2023languagebind} is a model that unifies multiple modalities (e.g., image, audio, video, depth) into a shared language-aligned embedding space.
    \item \textbf{SigLIP} \cite{zhai2023sigmoidlosslanguageimage} is a CLIP variant that replaces softmax with a sigmoid loss for more stable and scalable image–text contrastive learning.
    \item \textbf{MMMORRF} \cite{samuel2025mmmorrfmultimodalmultilingualmodularized} is a modular, multilingual video retrieval system that fuses visual, audio, and text (OCR/ASR) features through a modality-aware reciprocal rank fusion technique.
\end{itemize}

\begin{table}[]
\caption{Zero-shot video retrieval results of VLMs on the MSR-VTT benchmark. }
\resizebox{0.9\columnwidth}{!}{%
\begin{tabular}{l|ccc}
\hline \hline
\textbf{Model}                                        & \textbf{R@1} & \textbf{R@5} & \textbf{R@10} \\ \hline
Qwen2-VL-7B-Instruct                               & 0.002                       & 0.006                       & 0.010                         \\
GME-7B (Mean Pooling) & 0.411                       & 0.655                       & 0.764                        \\
GME-7B (Multi-image) & 0.201                       & 0.383                       & 0.485                        \\
LamRA & 0.447                       & 0.686                       & 0.786                        \\
InternVideo2-6B                                  & 0.559                       & 0.783                       & 0.851                        \\ \hline
Ours - Fine-tuned Model ($M_F$)  &   0.413 &   0.661 &  0.750\\
Ours - w. Retrieval Vector ($M_R$)                      & 0.476                       & 0.720                        & 0.798 \\ \hline \hline        
\end{tabular}
}
\label{table:msrvtt}
\end{table}

\begin{figure*}[t]
  \centering
  \subfloat[Search Mode]{%
    \includegraphics[width=0.24\textwidth]{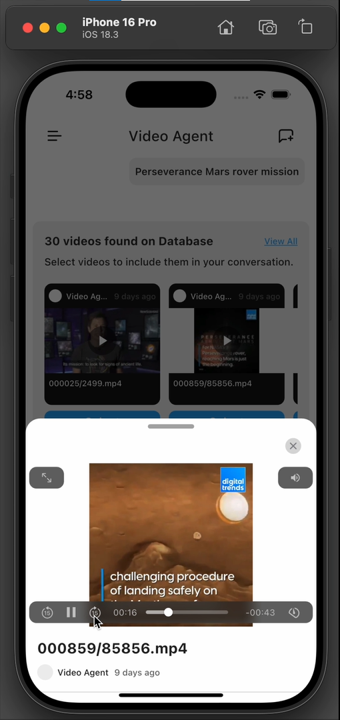}\label{fig:use1}}
  \hfill
  \subfloat[Agent Mode-Finding relevant videos and corresponding video summaries]{%
    \includegraphics[width=0.24\textwidth]{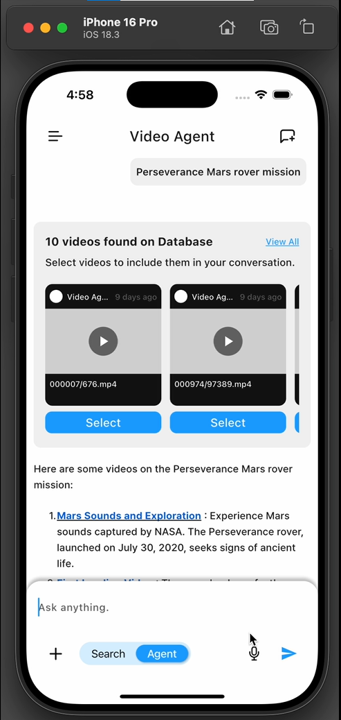}\label{fig:use2}}
  \hfill
  \subfloat[Agent Mode-Question answering (User Questioning using Multiple Videos)]{%
    \includegraphics[width=0.24\textwidth]{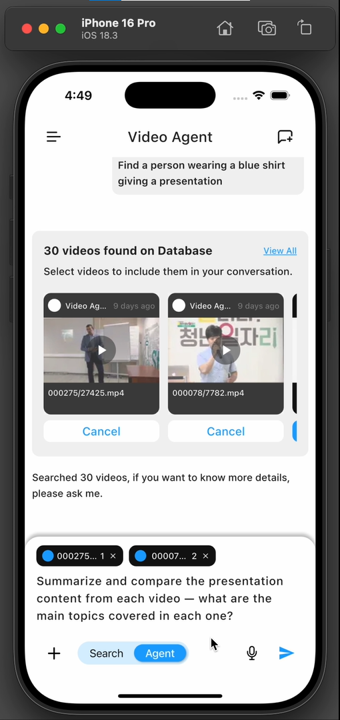}\label{fig:use3}}
  \hfill
  \subfloat[Agent Mode-Question answering (System Response)]{%
    \includegraphics[width=0.24\textwidth]{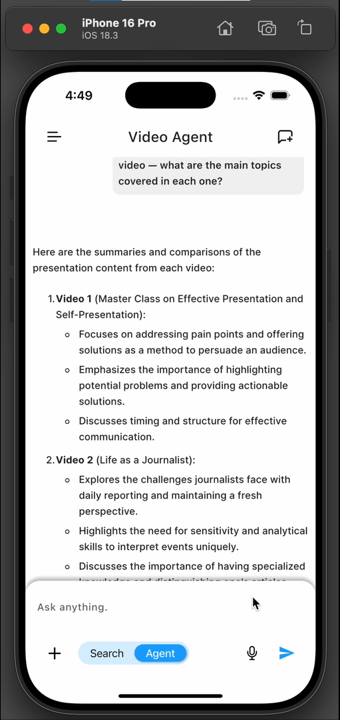}\label{fig:use4}}
  \caption{\textit{V-Agent} Use Case Examples.}
  \label{fig:usecases}
\end{figure*}

\begin{table}[]
\caption{Results of zero-shot video retrieval on the MultiVENT 2.0 benchmark. An asterisk (*) denotes results reported in prior work.}
\resizebox{0.8\columnwidth}{!}{%
\begin{tabular}{l|cc}
\hline \hline
\textbf{Model} & \textbf{nDCG@10} & \textbf{R@10} \\ \hline
\textbf{\textit{V-Agent}} (Ours) & \textbf{0.680}   & \textbf{0.676}  \\ \hline
InternVideo2-6B* \qquad\qquad & 0.005 & 0.004 \\ 
VAST* & 0.116 & 0.118 \\ 
CLIP* & 0.304 & 0.333 \\ 
LanguageBind* & 0.324 & 0.355 \\
SigLIP*& 0.375 & 0.409 \\
MMMORRF* & 0.586 & 0.611 \\ 

\hline \hline        
\end{tabular}
}
\label{table:multivent}
\end{table}

\section{Results and Analysis}
\subsection{MSR-VTT Evaluation Results}
In MSR-VTT, we aim to assess our retrieval models' effectiveness exclusively compared to other VLMs, without assistance from other AI agents. Table~\ref{table:msrvtt} shows the zero-shot evaluation performance of VLMs on the benchmark. Qwen2-VL-7B-Instruct exhibits low recall, suggesting that prompting alone is insufficient for general instruction-tuned VLMs to produce high-quality embeddings without a retrieval-specific training phase. Since GME models are not trained to process multiple images or videos, we evaluate with two different inference strategies. Mean pooling over embeddings of frame-level embeddings yields competitive performance, whereas multi-image inference performs poorly. Considering that frame-wise embedding extraction introduces more latency than a single multi-image pass, our models offer a favorable balance of performance and efficiency.

\subsection{Zero-shot Results on MultiVENT 2.0}
As shown in Table~\ref{table:multivent}, \textit{V-Agent} outperforms existing models on both nDCG@10 and Recall@10. The previous state-of-the-art model, MMMORRF, is similar to ours but incorporates SigLIP \cite{zhai2023sigmoidlosslanguageimage} and multilingual ColBERT-X \cite{neuralapproachestomultilingual, transferlearningapproaches} for embedding, whereas our framework relies on a single fine-tuned VLM and embeds both modalities into a shared embedding space.
On the contrary to the results in MSR-VTT (Table~\ref{table:msrvtt}), InternVideo2's performance on MultiVENT 2.0 is inferior compared to other models. \citet{samuel2025mmmorrfmultimodalmultilingualmodularized} reported that it may be due to its dependence on low quality captions and the complexity of queries in MultiVENT 2.0. We believe that engaging VLMs trained on multilingual datasets improves the understanding of queries across different languages, compared to baseline models that primarily emphasize modality alignment.

\begin{table}[]
\caption{Ablation test on MultiVENT 2.0. “Ret. Vec.” and “Desc.” are abbreviations for “Retrieval Vector” and “Description”, respectively.}
\resizebox{\columnwidth}{!}{%
\begin{tabular}{l|cccc|cc}
\hline \hline
Model                                                                                      & \multicolumn{1}{c}{\begin{tabular}[c]{@{}c@{}}\# of \\      frames\end{tabular}} & \multicolumn{1}{c}{\begin{tabular}[c]{@{}c@{}}Ret.\\      Vector\end{tabular}} & \multicolumn{1}{c}{Desc.} & \multicolumn{1}{c|}{\begin{tabular}[c]{@{}c@{}}Re-\\      ranking\end{tabular}}& nDCG@10 & R@10 \\ \hline
\textbf{\textit{V-Agent}} & 48 & O & O & O & \textbf{0.680}   & \textbf{0.676}     \\ 
 & 32 & O & O & O & 0.675 & 0.671    \\ \hline

\multirow{3}{*}{\begin{tabular}[c]{@{}l@{}}Frame \\      Difference\end{tabular}} & 48 & O & O & X    & 0.614   & 0.676     \\
& 32  & O    & O & X  & 0.611   & 0.671     \\
& 16   & O   & O  & X & 0.607   & 0.671     \\ \hline
\multirow{3}{*}{\begin{tabular}[c]{@{}l@{}}Ret. Vec. \\ \& Desc.\end{tabular}} & 16 & X & O& X & 0.597   & 0.66      \\
& 16 & O & X    & X  & 0.518   & 0.587     \\
& 16  & X  & X & X & 0.509   & 0.573  \\
\hline
\hline               

\end{tabular}
}
\label{table:multivent-ablation}
\end{table}

\subsection{Ablation}
Table \ref{table:multivent-ablation} presents the ablation results of \textit{V-Agent}, comparing different settings with respect to the number of frames, retrieval vector, description, and re-ranking.

\subsubsection{Impact of Re-ranking and Retrieval Vector.} 
\textit{V-Agent} demonstrates superior performance compared to models without re-ranking, highlighting the effectiveness of LLM-based re-ranking. Under the same frame settings, re-ranking improves performance by 6 percentage points in nDCG@10, which indicates that it significantly enhances the ranking of candidates according to their relevance to queries. We also observe a substantial improvement in video retrieval performance when the retrieval vector $\tau$ is incorporated into the weights of the fine-tuned model $M_R$. In the 16-frame setting, nDCG@10 and Recall@10 increase by 1 percentage point with the addition of the retrieval vector.

\subsubsection{Analysis on Frame Information and Description.}
To analyze whether the number of frames sampled from each video influences the retrieval performance, we evaluate our retrieval model using different frame settings (16, 32, and 48 frames). Although increasing the number of input frames enhances the model’s visual understanding, the resulting improvement in retrieval performance is relatively modest. In contrast, there is a significant drop in performance when video descriptions are omitted, showing that descriptions are a valuable resource for building a robust video chat system. Based on various experimental results, we validate the proposed \textit{V-Agent} pipeline and identify the most suitable configuration for an advanced video search system.

\section{Use Cases}
\textit{V-Agent} can be used for simple video search or agent-based question answering as illustrated in Figure~\ref{fig:vap}. Additional usage examples are shown in Figure~\ref{fig:usecases}. For instance, with the query "Perseverance Mars rover mission", users can choose between `search' and `agent' modes. In search mode (Figure~\ref{fig:use1}), the system quickly returns a list of relevant videos. 

In agent mode (Figures~\ref{fig:use2}, ~\ref{fig:use3}, ~\ref{fig:use4}), the system returns relevant videos together with concise summaries of their content. This mode also supports question answering over the retrieved results. As illustrated in Figure~\ref{fig:use3}, users can select multiple videos of interest and pose questions about their content. The chat agent then generates answers by integrating information from both the selected videos and the user query (Figure~\ref{fig:use4}), thereby offering an interactive and efficient video-based dialogue experience.

\section{Limitaions and Future Work}
In the current design of \textit{V-Agent}, visual information is not sufficiently incorporated during the re-ranking phase, which may constrain overall system performance. We hypothesize that integrating visual cues could yield more effective re-ranking outcomes and plan to explore this direction in future work. While the agent-based flow introduces higher latency compared to simple retrieval, we seek to mitigate this issue by streaming LLM responses to improve user experience. Our ultimate objective is to establish an optimal balance between retrieval effectiveness and system latency.

\section{Conclusion}
In this paper, we present \textit{V-Agent}, an interactive chat system based on three distinct AI agents that enables effective video search and question answering.  For effective video retrieval, we propose an efficient fine-tuning method that adapts VLMs into video-text retrieval models. The resulting VLM-based retrieval model is utilized in both the search agent and data indexing pipeline to generate visual and textual embeddings. Our approach addresses shortcomings of traditional text-based retrieval systems, allowing advanced interpretation of both visual and spoken content for comprehensive video understanding. Extensive experiments on the MultiVENT 2.0 benchmark demonstrate the effectiveness and robustness of our system. We believe that \textit{V-Agent} lays the groundwork for future advancements in interactive, multimodal video information access. 

\section*{Acknowledgments}
This work was supported by the Institute of Information \& Communications Technology Planning \& Evaluation (IITP) grant funded by the Korea Government (MSIT) (No. RS-2024-00338140, Development of learning and utilization technology to reflect sustainability of generative language models and up-to-dateness over time)

\section*{GenAI Usage Disclosure}
We employed GPT APIs (\texttt{gpt-4.1-mini}, \texttt{gpt-4o-mini}, and \texttt{gpt-4o}) to translate audio transcriptions into English and to develop the agents. The specific API versions used in each process are provided in footnotes throughout the paper. GenAI tools were also used during the writing of the paper, but the authors are fully accountable for the content.

\bibliographystyle{ACM-Reference-Format}
\bibliography{main}


\end{document}